\documentclass[twoside,twocolumn,10pt]{article}

\usepackage{fancyhdr}
\usepackage{geometry}
\usepackage{abstract}
\usepackage{graphicx}
\usepackage{titlesec}
\usepackage{ragged2e}
\usepackage{amssymb}
\usepackage{parskip}
\usepackage{amsmath}
\usepackage{booktabs}
\usepackage{array}
\usepackage{balance}
\usepackage[font=footnotesize,labelfont=bf,justification=raggedright,format=plain]{caption}
\usepackage{float}
\usepackage{xcolor}
\usepackage{hyperref}
\hypersetup{colorlinks=true, urlcolor=blue, linkcolor=blue, citecolor=blue}

\usepackage{enumitem}     % for itemize/enumerate [leftmargin=*]
\usepackage{subcaption}   % for the appendix sample grids

\usepackage{iftex}
\ifXeTeX
\else
  \PackageError{main}{This document requires XeLaTeX}
    {Use xelatex main.tex or latexmk -xelatex main.tex.}
\fi
\usepackage{fontspec}
\usepackage{polyglossia}
\setmainlanguage{english}
\setotherlanguage{arabic}
\newcommand{\gulmargfontpath}{./}
\IfFileExists{GulmargNastaleeq.ttf}{}{%
  \renewcommand{\gulmargfontpath}{Koshur_Pixel_v11/}%
}
\newfontfamily\kashfont[
  Path=\gulmargfontpath,
  Extension=.ttf,
  Script=Arabic,
  Scale=1.15
]{GulmargNastaleeq}
\newcommand{\ks}[1]{\mbox{\textarabic{\kashfont #1}}}

\newcommand{\ksfallback}[1]{\texttt{U+#1}}

\ExplSyntaxOn
\cs_if_exist:NF \tbl_save_outer_table_cols: { \cs_new:Npn \tbl_save_outer_table_cols: {} }
\ExplSyntaxOff

\usepackage[backend=biber,style=ieee,autolang=none]{biblatex}
\defbibheading{bibliography}[\refname]{%
  \section{\MakeUppercase{REFERENCES}}}

\newcolumntype{L}[1]{>{\raggedright\arraybackslash}p{#1}}
\newcolumntype{C}[1]{>{\centering\arraybackslash}p{#1}}
\newcolumntype{R}[1]{>{\raggedleft\arraybackslash}p{#1}}

\fancypagestyle{firstpage}{
  \fancyhf{}
  \fancyhead[LO]{\small }
 \fancyhead[RO]{\small}
  \fancyfoot[L]{\footnotesize \textsuperscript{*}Corresponding Author: Haq Nawaz Malik\\HF = Hugging Face; Dataset: https://huggingface.co/datasets/Omarrran/Koshur\_Pixel}
  \fancyfoot[R]{\footnotesize \textcopyright\ 2026\\ \hspace{0.5cm} }

}

\titleformat{\section}{\normalsize\bfseries\uppercase}{\thesection.}{1em}{}
\titleformat{\subsection}{\normalsize\bfseries\flushleft}{\thesubsection.}{1em}{}
\titleformat{\subsubsection}{\normalsize\bfseries\itshape\flushleft}{\thesubsubsection.}{1em}{}

\begin{document}

\twocolumn[
\begin{@twocolumnfalse}
\sloppy

\begin{center}
    \vspace*{0.5em}
    {\Large\bfseries Koshur Pixel: A Large-Scale Synthetic OCR Dataset for Kashmiri\par}

    \vspace{1.5em}

    \small
    \begin{tabular*}{\textwidth}{@{\extracolsep{\fill}}ccc@{}}
        \textbf{Haq Nawaz Malik*} & \textbf{Faizan Iqbal} & \textbf{Nahfid Nissar} \\
        ORCID: 0009-0003-1994-7640 & ORCID: 0009-0002-8998-9347 & ORCID: 0009-0002-2805-4687 \\
        HF: @Omarrran  & HF: @faizaniqbal & HF: @nafiboi
    \end{tabular*}

    \vspace{2em}
\end{center}
\thispagestyle{firstpage}

\section*{\hspace{-1.5mm}Abstract}
\fontsize{10}{12}\selectfont
\justifying
\emergencystretch=6em
\noindent
Optical Character Recognition (OCR) for low-resource languages remains one of the most significant hurdles in achieving global digital equity and linguistic inclusivity. Kashmiri, an Indo-Aryan language spoken by approximately 7 million people, currently exists in a profound ``digital void'' where mainstream vision-language models demonstrably fail to process its unique Perso-Arabic Nastaliq script. This systemic failure is primarily attributed to two compounding factors: the extreme visual and orthographic complexity inherent in Nastaliq script, characterized by intricate contextual letter morphing, highly non-linear glyph layering with complex ligatures and a critical, pervasive absence of large-scale, accurately annotated training data. Traditional manual annotation for such a script is prohibitively expensive, time-consuming, and prone to human error, creating an intractable ``data deadlock.'' In this paper, we introduce \textbf{Koshur Pixel}, a comprehensive, meticulously engineered synthetic OCR dataset specifically designed to address the challenges of the Kashmiri language. Comprising 613,078 high-fidelity image-text pairs, Koshur Pixel represents the first substantial effort to provide a robust training resource for Kashmiri OCR. Our methodology leverages the \textit{SynthOCR-Gen} pipeline \cite{malik2026synthocr} to programmatically render the \textit{KS-PRET-5M} text corpus \cite{malik2026kspret5m} into multi-font, multi-granularity images. These images span a spectrum from isolated words to full-page layouts, meticulously preserving the linguistic nuances of Nastaliq. Furthermore, our pipeline incorporates an extensive suite of 25+ advanced data augmentation techniques, simulating many real-world document degradations, including subtle ink bleed, realistic paper aging, scanner artifacts, and various geometric distortions. By effectively breaking the long-standing data barrier, Koshur Pixel establishes a foundational, openly accessible resource for fine-tuning state-of-the-art OCR systems, thereby enabling the critical digitization of Kashmiri historical archives, fostering the development of essential accessibility tools, and ultimately ensuring the digital survival and thriving of the Kashmiri language in the 21st century. We provide a detailed analysis of the script's unique visual and orthographic challenges and rigorously demonstrate how programmatic data generation offers a scalable and cost-effective solution to bypass the prohibitive expenses associated with manual annotation in severely low-resource linguistic contexts.

\vspace{0.3cm}
\noindent\textbf{Keywords:}
{\small Optical Character Recognition (OCR), Kashmiri Language, Nastaliq Script, Synthetic Dataset Generation, Low-Resource NLP, Digital Preservation, Computer Vision, Document Image Analysis, Linguistic Diversity, Machine Learning Datasets.\par}

\vspace{0.8cm}
\fussy
\end{@twocolumnfalse}
]

% -----------------------------
% Main Text (content unchanged, one section per file)
% -----------------------------

\section{Introduction: The Digital Void and Linguistic Inequity}\label{sec:introduction}

The rapid and pervasive advancement of artificial intelligence (AI) and digital technologies has fundamentally reshaped human interaction with information, commerce, and culture. Yet, this transformative progress has not been uniformly distributed across the global linguistic landscape. A significant disparity persists, creating a significant divide between technologically privileged, high-resource languages and their underserved, low-resource counterparts. While dominant languages such as English, Mandarin, and Spanish benefit from a rich ecosystem of sophisticated digital tools that encompass real-time machine translation, advanced search engines, and highly accurate voice assistants, thousands of other languages face an existential threat of "digital extinction" or "digital death" \cite{kashmiri_unesco}. This phenomenon describes a scenario where a language, despite having a vibrant community of speakers, gradually loses its functional relevance in the digital sphere due to a profound lack of technological support.

Kashmiri, an Indo-Aryan language belonging to the Dardic sub-branch, exemplifies this critical issue. Spoken by an estimated 7 million people primarily across the Kashmir Valley in India and its neighbouring parts, Kashmiri boasts a rich and ancient literary tradition, with documented history spanning over a millennium. However, in the contemporary digital world, Kashmiri remains largely invisible to machines. This invisibility is not a consequence of a diminishing speaker base or a lack of cultural vitality; rather, it stems from a profound technical deficit, most notably the absence of robust Optical Character Recognition (OCR) systems for its unique Perso-Arabic Nastaliq script.

Optical Character Recognition is a cornerstone technology of the modern digital age, serving as the essential bridge that converts visual representations of text (from scanned documents, photographs, or handwritten notes) into machine-readable, editable, and searchable digital data. For high-resource languages, OCR is considered a mature and largely "solved" problem, with commercial and open-source systems achieving near-human levels of accuracy on both printed and even complex handwritten texts. In  contrast, for Kashmiri, production-ready OCR capabilities are virtually non-existent. This critical technological gap means that vast repositories of Kashmiri literature, historical manuscripts, ancient administrative records, and contemporary journalistic output remain inaccessible. They are effectively locked within physical archives, impervious to modern computational tools that drive academic research, facilitate educational content creation, and underpin efficient governance and public services.

This pervasive lack of digital infrastructure has profound socio-technical implications, contributing directly to the UNESCO classification of Kashmiri as a vulnerable language \cite{kashmiri_unesco}. The vulnerability arises not from a decline in oral transmission, but from its systemic displacement by Urdu and English in digital communication. Without functional OCR, Kashmiri speakers are often compelled to interact with digital platforms and information systems using non-native languages, thereby marginalizing their linguistic heritage in the very spaces that define modern life. The inability to digitize, search, and process Kashmiri text computationally hinders scholarly research, impedes the development of educational materials, and severely limits the language's presence and utility on the internet.

The core challenge in developing a functional OCR system for Kashmiri is multifaceted, arising from two primary, interconnected problems. Firstly, the inherent visual and orthographic complexity of the Perso-Arabic Nastaliq script presents formidable technical hurdles. Unlike the relatively linear, discrete, and left-to-right nature of Latin-based scripts, Nastaliq is a highly cursive, right-to-left writing system characterized by an intricate interplay of contextual letter morphing, non-linear stacking, and the pervasive use of complex ligatures. These features defy the fundamental assumptions embedded in most mainstream OCR algorithms, which were primarily designed for simpler, non-cursive scripts. Secondly, and perhaps more critically, there is a complete and systemic absence of large-scale, accurately annotated training data specifically for Kashmiri Nastaliq. Modern deep learning models, including advanced Convolutional Recurrent Neural Networks (CRNNs) and sophisticated Vision Transformers (ViTs), demand vast quantities of meticulously labeled data, often hundreds of thousands to millions of samples, to achieve robust performance. For a language like Kashmiri, which historically lacks significant institutional funding, dedicated research infrastructure, and a large pool of computational linguists, the manual annotation of such a dataset is not merely challenging; it is practically impossible. This creates an intractable "data deadlock": models cannot be effectively trained due to a lack of data, and building training data from old printed documents, manuscripts, and archives also requires exact Unicode reference transcriptions for each physical source. Since such aligned digital text is largely unavailable, these materials cannot be readily converted into supervised OCR training examples.

In this seminal work, we directly confront and resolve this long-standing data deadlock by introducing \textbf{Koshur Pixel}, the first large-scale, high-fidelity synthetic OCR dataset specifically engineered for the Kashmiri language. Our innovative approach is grounded in the philosophy of "programmatic data generation," which fundamentally inverts the traditional, labor-intensive data collection paradigm. Instead of embarking on the arduous and error-prone task of manually labeling noisy real-world scans, we leverage the advanced \textit{SynthOCR-Gen} pipeline \cite{malik2026synthocr} to programmatically render a deeply cleaned and linguistically validated Kashmiri text corpus (the \textit{KS-PRET-5M} corpus \cite{malik2026kspret5m}) into high-quality synthetic images. This process ensures perfect ground-truth alignment, a critical advantage over human-annotated datasets, and allows for the scalable generation of over half a million samples, a quantity more than sufficient to effectively train and fine-tune state-of-the-art OCR models.

The overarching contributions of this research are multi-faceted and aim to address both the technical and socio-linguistic dimensions of the problem:
\begin{itemize}[leftmargin=*]
    \item \textbf{A Pioneering Large-Scale Dataset:} We introduce and publicly release Koshur Pixel, a meticulously curated dataset comprising 613,078 high-quality image-text pairs. This dataset is structured across four distinct levels of granularity (individual words, complete sentences, multi-line paragraphs, and full-page layouts), thereby enabling the training of a diverse range of OCR models, from simple character classifiers to complex layout-aware document transformers.
    \item \textbf{In-Depth Linguistic and Orthographic Analysis:} We provide an exhaustive analysis of the unique visual, orthographic, and Unicode-specific challenges inherent in the Kashmiri Nastaliq script. This analysis systematically identifies the precise failure modes of existing, mainstream OCR systems when confronted with the intricacies of Kashmiri, laying the groundwork for targeted model development.
    \item \textbf{Robust and Reproducible Synthetic Generation Pipeline:} We detail the architecture and implementation of a sophisticated synthetic data generation pipeline. This pipeline integrates advanced multi-font rendering techniques, powered by the browser's native Canvas-based text-shaping engine, and a comprehensive suite of 25+ diverse augmentation techniques meticulously designed to simulate the full spectrum of real-world document degradations and noise. This ensures that models trained on Koshur Pixel are robust and generalize effectively to authentic scanned documents.
    \item \textbf{Socio-Technical Impact and Ethical Considerations:} We engage in a critical discussion of the broader socio-technical implications of this work, particularly its role in language preservation and the ethical considerations surrounding AI development for vulnerable linguistic communities. We also highlight persistent technical barriers, such as the scarcity of high-performance computing resources in low-resource regions, which remain crucial for the full realization of digital linguistic equity.
\end{itemize}

By providing this unprecedented resource and detailing the robust methodology behind its creation, we aspire to ensure that the Kashmiri language is not merely preserved but actively thrives in the digital age. Koshur Pixel transcends being merely a dataset; it represents a deliberate and impactful technical intervention against the ongoing digital exclusion of 7 million speakers, offering a tangible pathway towards linguistic empowerment through artificial intelligence.

The remainder of this paper is structured as follows: Section \ref{sec:background} provides a comprehensive review of the historical context of Kashmiri Natural Language Processing (NLP) and the evolutionary trajectory of OCR technologies. Section \ref{sec:linguistics} delves into a deep linguistic and orthographic analysis of the Kashmiri Nastaliq script, detailing its unique complexities. Section \ref{sec:methodology_text_corpus} meticulously describes the multi-stage construction of the Koshur Pixel pipeline. Section \ref{sec:analysis} presents a detailed statistical and qualitative analysis of the generated dataset. Section \ref{sec:impact} discusses the broader socio-technical implications, use cases, and inherent limitations of our work. Finally, Section \ref{sec:conclusion} concludes the paper by summarizing our findings and outlining promising avenues for future research.

\section{Background and Related Work: Kashmiri NLP and OCR Evolution}\label{sec:background}

\subsection{The Historical Context of Kashmiri Digitization and NLP}
The journey of the Kashmiri language into the digital realm has been historically complex and fraught with significant technical hurdles. For several decades, the de facto standard for digital typesetting and printing of Kashmiri text was the proprietary desktop publishing software, \textit{InPage} \cite{inpage_wikipedia}. While InPage excelled at rendering the intricate Nastaliq script with high calligraphic fidelity, its reliance on a non-Unicode, custom encoding scheme created a massive and enduring problem for Natural Language Processing (NLP) and general digital accessibility. Text created in InPage was effectively "locked" within its ecosystem, rendering it unreadable and unprocessable by standard NLP tools, search engines, or modern computational linguistics frameworks. This led to the creation of a substantial "lost generation" of digital Kashmiri text, comprising millions of words across countless books, newspapers, and official documents that existed in digital file formats but were linguistically inaccessible to machines.

Recent pioneering efforts have focused on "unlocking" this invaluable textual heritage. Notably, Malik \cite{malik2024inpage} developed a robust and highly accurate InPage-to-Unicode converter, achieving an impressive 98.7\% conversion accuracy. This critical technological breakthrough enabled the recovery and standardization of millions of words of previously inaccessible Kashmiri literature. The output of this conversion process formed the foundational \textit{KS-PRET-5M} corpus \cite{malik2026kspret5m}, which is the primary textual source for the Koshur Pixel dataset. Without these prior, arduous efforts in text recovery and standardization, the construction of a large-scale, linguistically pure OCR dataset for Kashmiri would have remained an insurmountable challenge.

Beyond text recovery, general Kashmiri NLP research remains nascent. Efforts have been made in areas such as morphological analysis and part-of-speech tagging, often relying on rule-based systems or small, manually annotated corpora. The development of large-scale language models for Kashmiri, akin to those for high-resource languages, has been severely hampered by the scarcity of digitized text and the complexities of its script. The \textit{KS-PRET-5M} corpus, therefore, represents a significant leap forward, providing a much-needed resource for pretraining modern language models, as demonstrated by early work on byte-level language modeling for Kashmiri using Transformer architectures [11].

\subsection{Evolution of OCR Architectures}

Optical Character Recognition (OCR) has evolved through four major architectural paradigms. Early systems relied on template matching and handcrafted feature engineering, performing adequately on clean machine-printed text but failing under variations in font, style, and image quality \cite{mori1992ocrreview}. The introduction of statistical learning methods, particularly Hidden Markov Models (HMMs), improved robustness by modeling sequential character dependencies, although accurate character segmentation remained a critical bottleneck \cite{kuo1993hmmocr}. The deep learning era transformed OCR through Convolutional Recurrent Neural Networks (CRNNs) and Connectionist Temporal Classification (CTC), enabling end-to-end recognition without explicit segmentation and substantially improving performance on cursive and complex scripts \cite{graves2006ctc,shi2015crnn}. More recently, Transformer-based vision-language models such as TrOCR and Donut have reframed OCR as an image-to-text sequence generation task, leveraging large-scale pretraining to achieve state-of-the-art results \cite{trocr2021,donut2022}. However, these models remain constrained in low-resource settings, where limited exposure to underrepresented scripts, orthographic variations, and diacritic-rich writing systems significantly degrades recognition accuracy. Consequently, low-resource languages such as Kashmiri continue to require specialized datasets and script-aware OCR architectures to bridge the performance gap.

\subsection{The Role of Synthetic Data in Computer Vision and OCR}
The strategic use of synthetic data has emerged as a powerful paradigm in computer vision, particularly in scenarios where real-world annotated data is scarce, expensive, or difficult to obtain. Synthetic data generation involves creating artificial datasets that mimic the characteristics of real data, often with perfect ground-truth labels. This approach offers several compelling advantages:
\begin{itemize}[leftmargin=*]
    \item \textbf{Scalability:} Synthetic data can be generated in virtually unlimited quantities, overcoming the bottleneck of manual annotation.
    \item \textbf{Perfect Ground Truth:} By controlling the generation process, synthetic data inherently possesses accurate labels, eliminating human annotation errors.
    \item \textbf{Diversity Control:} Specific variations (e.g., fonts, backgrounds, degradations) can be systematically introduced to enhance model robustness.
    \item \textbf{Cost-Effectiveness:} Generating synthetic data is often significantly cheaper and faster than collecting and annotating real-world data.
\end{itemize}

In the context of OCR, pioneering projects like \textit{SynthText} \cite{synthtext} demonstrated the efficacy of training models on synthetic text rendered onto natural images, showing that such models could generalize remarkably well to real-world scenes. However, the majority of synthetic OCR research and available tools have historically focused on Latin, Cyrillic, or CJK (Chinese, Japanese, Korean) scripts. The unique cursive, contextual, and multi-layered nature of Nastaliq presents a far greater challenge for synthetic rendering. Naive rendering approaches often fail to correctly form ligatures, misplace diacritics, or incorrectly handle the right-to-left flow, leading to synthetically generated images that are linguistically inaccurate and visually unrepresentative of true Nastaliq. Our work, leveraging the \textit{SynthOCR-Gen} pipeline \cite{malik2026synthocr}, specifically addresses these complex rendering challenges, providing a high-fidelity solution for RTL cursive scripts.

\section{Deep Linguistic Analysis of the Kashmiri Nastaliq Script}\label{sec:linguistics}

Kashmiri is predominantly written in the Perso-Arabic script, specifically adopting the Nastaliq calligraphic style, which accounts for over 98.5\% of its written material. Nastaliq, a term derived from the Persian words \textit{naskh} (copying) and \textit{ta'liq} (hanging), is renowned for its aesthetic beauty and fluid, cursive nature. However, these very characteristics that make it visually appealing also render it one of the most challenging scripts for Optical Character Recognition (OCR) systems, particularly those not explicitly designed for its unique orthographic rules.

\subsection{The "Nastaliq Void": Intrinsic Visual Complexity}
The fundamental distinction of Nastaliq from other Perso-Arabic scripts, such as Naskh (commonly used for Arabic), lies in its highly cursive and non-linear structure. This intrinsic visual complexity creates a significant "Nastaliq Void" for conventional OCR algorithms. Key characteristics contributing to this complexity include:

\subsubsection{Slanted Baseline and Vertical Stacking}
Unlike Latin or Naskh scripts where characters typically align along a horizontal baseline, words in Nastaliq exhibit a characteristic downward slope, flowing diagonally from the top-right to the bottom-left. This creates a dynamic, irregular baseline within each word and across lines of text. Furthermore, characters and their associated diacritics (such as vowel marks) are not simply placed side-by-side; they are frequently stacked vertically, sometimes forming two or three layers within the same word unit. This vertical compression and non-linear arrangement pose severe challenges for traditional line and word segmentation algorithms, which often assume a predominantly horizontal text flow. A model expecting a flat baseline will misinterpret the vertical shifts as noise or segmentation errors, leading to incorrect bounding box detection and character isolation.

\subsubsection{Contextual Letter Morphing and Complex Ligatures}
In the Perso-Arabic writing system, most letters possess four distinct contextual forms: isolated, initial, medial, and final. The visual representation of a letter changes dramatically based on its position within a word and its connection to adjacent characters. For instance, the letter \textit{Hey} (\ks{ہ}) undergoes significant morphological changes depending on whether it appears at the beginning, middle, or end of a word. In Kashmiri Nastaliq, this contextual morphing is further compounded by the pervasive use of complex ligatures. Ligatures are indivisible graphical units formed by the fusion of two or more characters into a single, aesthetically integrated glyph. A single visual "blob" in a Nastaliq word can represent a sequence of multiple Unicode characters. For example, the combination of \ks{لام} (lam) and \ks{الف} (alif) often forms a distinct ligature \ks{لا}. Recognizing these ligatures requires sophisticated text shaping engines that understand the underlying character sequences, rather than treating each visual component as an independent character. OCR systems that fail to account for these ligatures often missegment them into individual, incorrect characters or fail to recognize the sequence altogether, leading to severe transcription errors.

\subsection{The Challenge of Unique Kashmiri Diacritics and Codepoints}
While Kashmiri shares a common script base with Urdu and Persian, it employs a modified version of the Perso-Arabic alphabet that includes several unique diacritics and characters. These specialized codepoints are crucial for accurately representing the distinct phonological inventory of the Kashmiri language, particularly its rich vowel system and aspirated consonants, which do not have direct equivalents in standard Arabic or Urdu. Table \ref{tab:unicode_kashmiri} provides a list of some of the most critical unique Kashmiri Unicode codepoints.

\begin{table}[htbp]
\caption{Critical Unique Kashmiri Unicode Codepoints and their Challenges for OCR}
\label{tab:unicode_kashmiri}
\centering
\small
\setlength{\tabcolsep}{4pt}
\resizebox{\columnwidth}{!}{%
\begin{tabular}{lll}
\toprule
\textbf{Character} & \textbf{Name/Description} & \textbf{Unicode Value} \\
\midrule
\ks{ٲ} & Alef with Wasla & U+0672 \\
\ks{ۆ} & Waw with Above Dot & U+06C6 \\
\ksfallback{06C4} & Waw with Ring & U+06C4 \\
\ks{ٕ} & Subscript Alef & U+0656 \\
\midrule
\multicolumn{3}{l}{\textbf{Composite Examples:}}\\
\ks{ہٕ} & He with Subscript Alef & U+06C1 U+0656 \\
\ks{چھ} & Aspirated Ch (Cheh with Heh Doachashmee) & U+0686 U+06BE \\
\bottomrule
\end{tabular}}
\vspace{-0.02cm}
\caption*{Note: The Subscript Alef (U+0656) is a combining character that attaches to a base letter, often leading to vertical stacking issues. Composite characters like \ks{ہٕ} are particularly problematic as they are often misidentified as standard Arabic/Urdu characters or fragmented during recognition.}
\end{table}

Standard OCR models, typically pretrained on Arabic or Urdu corpora, frequently misinterpret or entirely ignore these unique Kashmiri marks. This leads to several common failure modes:
\begin{enumerate}[leftmargin=*]
    \item \textbf{Noise Filtering:} The diacritics are often treated as extraneous visual noise, such as smudges or scanner artifacts, and are consequently filtered out or ignored during preprocessing, leading to a complete loss of phonological information.
    \item \textbf{Character Substitution:} The model may incorrectly substitute a unique Kashmiri character with the closest visually similar Arabic or Urdu character it has been trained on. This results in a transcription that is orthographically incorrect and semantically altered, rendering the text linguistically meaningless.
    \item \textbf{Segmentation Failure:} Due to the vertical stacking nature of many diacritics, OCR systems often struggle to correctly segment the base character from its diacritical mark. This can lead to fragmented character recognition or incorrect bounding box predictions, further degrading accuracy.
\end{enumerate}

A robust Kashmiri OCR system must therefore be explicitly trained on data that accurately represents these unique characters in their correct Nastaliq forms, ensuring that their presence and position are correctly interpreted rather than discarded as noise.

\subsection{Right-to-Left (RTL) and Bidirectional Text Processing}
Kashmiri, like other Perso-Arabic languages, is written and read from right-to-left (RTL). While modern text rendering engines are capable of handling RTL scripts, many OCR pipelines, particularly those developed with a primary focus on Latin-based languages, are optimized for left-to-right (LTR) reading orders. This fundamental directional mismatch can lead to significant errors:
\begin{itemize}[leftmargin=*]
    \item \textbf{Reversed Word Order:} If the OCR system incorrectly assumes an LTR reading order, it may output words in reverse sequence, rendering the transcribed text unintelligible.
    \item \textbf{Bidirectional Text (Bidi) Complexity:} Kashmiri text often incorporates numerals (which are LTR) or embedded English words. Correctly handling these bidirectional segments, where the reading direction shifts within a single line, is a complex task. A failure in Bidi processing can result in jumbled character sequences or incorrect logical ordering of text segments, making the output unusable for downstream NLP tasks.
\end{itemize}

These linguistic and orthographic intricacies collectively define the "Nastaliq Void" that current general-purpose OCR systems fall into when confronted with Kashmiri. Addressing these challenges requires a data generation strategy that is deeply informed by the script's unique properties, ensuring that the synthetic data accurately reflects the visual and linguistic realities of Kashmiri.

\section{Methodology I: The KS-PRET-5M Text Corpus Pipeline}\label{sec:methodology_text_corpus}

The efficacy and linguistic fidelity of any synthetic OCR dataset are fundamentally contingent upon the quality and representativeness of its underlying text corpus. For the Koshur Pixel dataset, the foundation is the \textit{KS-PRET-5M} corpus \cite{malik2026kspret5m}, which stands as the most extensive and meticulously cleaned collection of Kashmiri text currently available. The construction of this corpus involved a rigorous, multi-stage pipeline designed to transform raw, often noisy, digitized text into a "gold standard" source suitable for high-fidelity image rendering. This pipeline comprises 11 distinct stages, each addressing specific challenges inherent in processing low-resource language data from diverse sources.

\subsection{Corpus Acquisition and Normalization}

The textual foundation \textit{KS-PRET-5M} corpus \cite{malik2026kspret5m} was assembled from diverse digitized sources, including historical archives, literary texts, news articles, and web-crawled content. Owing to substantial heterogeneity in encoding standards, orthographic conventions, and document quality, the raw corpus underwent an extensive eleven-stage normalization pipeline designed to maximize linguistic consistency and data quality. The pipeline included encoding repair and mojibake correction, removal of markup artifacts, scrubbing of personally identifiable information, script-purity filtering, Unicode NFC normalization, whitespace and punctuation standardization, correction of line-break and hyphenation errors, sentence- and paragraph-level deduplication, length-based filtering, and final linguistic validation through manual review. Collectively, these procedures eliminated noise, corrected encoding inconsistencies, standardized textual representations, and ensured adherence to Kashmiri orthographic norms. The resulting corpus comprises 5.09 million words and provides a high-quality, linguistically reliable textual resource that serves as the foundation for subsequent synthetic image generation and OCR dataset construction.

\section{Methodology II: The SynthOCR-Gen Rendering Engine}\label{sec:methodology_rendering}

The second, and arguably most critical, phase in the construction of the Koshur Pixel dataset involves the high-fidelity rendering of the cleaned Unicode text from the \textit{KS-PRET-5M} corpus into visual images. This process is executed by the \textit{SynthOCR-Gen} pipeline \cite{malik2026synthocr}, a specialized, fully client-side and browser-based tool engineered to overcome the inherent complexities of rendering Right-to-Left (RTL) cursive scripts, particularly the Nastaliq style. Unlike simpler rendering approaches designed for Latin-based scripts, SynthOCR-Gen relies on the web platform's native, standards-compliant text-shaping and rendering capabilities to ensure the accurate and linguistically faithful visual representation of Kashmiri text.

\subsection{The Client-Side Rendering and Text-Shaping Pipeline}
A detailed technical description of the generator architecture, implementation workflow, and rendering controls can be found in the \textit{SynthOCR-Gen} pipeline paper \cite{malik2026synthocr}. Accurate rendering of Nastaliq is not a trivial task; it requires a text-shaping process capable of interpreting and applying the complex layout rules embedded within OpenType fonts. SynthOCR-Gen is implemented as a fully client-side, browser-based application and performs rendering entirely within the web platform's native graphics and typography stack, without dependence on any external native toolchain. Fonts are supplied to the runtime through the FontFace API; the input Unicode text is first segmented into its constituent grapheme clusters using the \texttt{Intl.Segmenter} API configured for Arabic-script locales, and is normalised to Unicode Normalization Form C (NFC) to guarantee a consistent character representation prior to rendering. The shaped text is then drawn onto an HTML5 Canvas~2D surface. When text is committed to the canvas, the browser's native text-shaping engine resolves the OpenType layout features encoded in the active font and emits the correct sequence of positioned glyphs. Delegating low-level shaping to the highly optimised, standards-compliant engine built into modern browsers ensures faithful rendering of complex scripts while keeping the tool lightweight and portable. The following OpenType layout features, applied automatically by this pipeline, are essential to the visual integrity of Nastaliq:
\begin{itemize}[leftmargin=*]
    \item \textbf{Glyph Substitution (GSUB):} This feature handles the contextual forms of characters. For example, a single Unicode character for 'Alef' (\ks{ا}) may possess distinct glyphs for its isolated, initial, medial, and final forms. The shaping engine selects the appropriate glyph based on each character's position within a word and its relationship to the surrounding characters.
    \item \textbf{Glyph Positioning (GPOS):} GPOS features dictate the precise placement of glyphs relative to one another. In Nastaliq, this is particularly vital for the correct positioning of diacritics (vowel marks and dots) above or below base characters, and for managing the vertical stacking of letters. Incorrect GPOS application leads to misaligned diacritics or overlapping characters, rendering the text unreadable.
    \item \textbf{Ligature Formation:} As discussed in Section \ref{sec:linguistics}, ligatures are fundamental to Nastaliq. The shaping engine identifies character sequences that should coalesce into a ligature (e.g., 'lam' + 'alif' $\rightarrow$ 'la') and substitutes them with the single, composite ligature glyph defined in the font. This ensures that the rendered output is calligraphically correct and visually coherent.
    \item \textbf{Right-to-Left (RTL) and Bidirectional (Bidi) Support:} The rendering context is explicitly configured for right-to-left flow, and the engine correctly reorders characters and segments so that the logical order of the text is preserved while adhering to the visual conventions of RTL scripts. This is essential for the correct treatment of embedded left-to-right runs, such as numerals, within Kashmiri text.
\end{itemize}

Once the glyph sequence has been shaped and positioned, the Canvas~2D rasteriser converts the vector glyph outlines of the selected font into high-quality, anti-aliased pixel data on the image surface, producing smooth and legible text across the full range of font sizes used during generation. Higher-level layout concerns, including text alignment, padding, line breaking, and the vertical arrangement of multi-line blocks, are governed by the application's own layout logic operating over the canvas. This separation between native glyph shaping and application-level layout is what enables the faithful generation of the multi-line paragraph and full-page samples described later in this section.

\subsection{Multi-Font Strategy for Enhanced Robustness}
To prevent models trained on Koshur Pixel from overfitting to the stylistic peculiarities of a single typeface, SynthOCR-Gen employs a multi-font rendering strategy. This approach ensures that the generated images exhibit sufficient visual diversity, compelling the OCR models to learn script-invariant features rather than font-specific patterns. We primarily utilize two  widely used high-quality, open-source Kashmiri  fonts:
\begin{itemize}[leftmargin=*]
    \item \textbf{Gulmarg Nastaleeq:} This font, distributed through Kashmiri language resource portals \cite{kashmiri_language_fonts}, represents a traditional, high-contrast Nastaliq style, commonly found in printed books, newspapers, and classical Kashmiri literature. Its intricate curves and pronounced diagonal flow are characteristic of historical Nastaliq calligraphy.
    \item \textbf{Afan Koshur Naksh:} This Unicode-compliant Kashmiri font was developed to address Kashmiri-specific Perso-Arabic shaping requirements, including positional glyph forms, GSUB/GPOS behavior, and Kashmiri-specific character support \cite{lawey2011afan}. It offers a more contemporary, 'Naksh-style' interpretation of Kashmiri script. While still retaining Nastaliq elements, it often presents a slightly more upright and less dramatically slanted appearance, making it prevalent in modern digital displays, web content, and social media. Its inclusion ensures coverage of evolving digital typographic trends.
\end{itemize}
By dynamically switching between these fonts during the rendering process, the dataset inherently captures a broader spectrum of visual variations, thereby enhancing the generalization capabilities of models trained on Koshur Pixel.

\subsection{Granularity Levels of Rendered Output}
SynthOCR-Gen is designed to produce synthetic images at four distinct levels of granularity, catering to a wide array of OCR research and application needs. This multi-granularity approach allows for flexible model training, from low-level character recognition to high-level document understanding:
\begin{itemize}[leftmargin=*]
    \item \textbf{Word Level:} Individual words are rendered as separate image files, typically at a fixed height (e.g., 64 pixels) with variable width. These samples are ideal for training character-level recognition models, such as those based on CRNN architectures, where the focus is on recognizing isolated word forms.
    \item \textbf{Sentence Level:} Complete sentences or single lines of text are rendered. This granularity captures the horizontal (or diagonal) flow of the script within a line, including inter-word spacing and the interaction of ligatures across word boundaries. It is particularly useful for training sequence-to-sequence models that process entire lines of text.
    \item \textbf{Paragraph Level:} Multi-line blocks of text, typically 3-5 lines, are rendered. This level introduces the challenge of vertical spacing between lines, line breaking, and the overall layout of a small text block. Models trained on paragraph-level data must learn to handle more complex spatial relationships between text lines.
    \item \textbf{Page Level:} Full-page layouts, simulating complete document pages with varying margins, multiple paragraphs, and potentially headers/footers, are generated. This is the most complex granularity, requiring models to perform comprehensive document analysis, including text detection, line segmentation, and reading order determination. Page-level samples are crucial for training end-to-end document understanding transformers like Donut [5] or Pix2Struct.
\end{itemize}

Each rendered image is accompanied by its precise ground-truth text, bounding box coordinates for words and characters, and metadata regarding the font, size, and augmentations applied. This rich annotation ensures that the Koshur Pixel dataset is not only visually diverse but also provides comprehensive information for advanced OCR model development.

\section{Methodology III: Advanced Visual Augmentation Suite}\label{sec:methodology_augmentation}

A critical challenge in the development of synthetic datasets for computer vision, particularly for OCR, is the "synthetic-to-real" gap. This phenomenon refers to the performance degradation observed when a model trained exclusively on clean, artificially generated images is deployed in real-world scenarios involving noisy, degraded, or variably formatted documents. To mitigate this gap and enhance the robustness and generalization capabilities of models trained on Koshur Pixel, our pipeline incorporates an extensive and meticulously designed suite of over 25 distinct visual augmentation techniques. These augmentations are applied stochastically during the image generation process, simulating a wide spectrum of real-world document degradations and environmental capture conditions.

Our augmentation strategy is broadly categorized into four groups: geometric transformations, photometric distortions, noise injection, and document-specific degradations. Each category addresses a particular type of variability or artifact commonly encountered in scanned or photographed documents.

\subsection{Geometric Transformations}
Geometric augmentations are applied to simulate variations in document alignment, perspective, and physical deformation that occur during scanning or photography. These transformations ensure that the OCR model becomes invariant to minor spatial distortions.

\subsubsection{Random Rotation}
Images are rotated by a small random angle, typically within $\pm 5^{\circ}$. This simulates slight misalignments when a document is placed on a scanner bed or photographed without perfect orientation. The rotation is performed around the center of the text bounding box to minimize cropping of text.

\subsubsection{Shear Transformation}
Shearing operations are applied along both the X and Y axes, introducing a slant to the text. This mimics the effects of perspective distortion or physical warping of paper. Shear factors are typically sampled from a uniform distribution within a small range (e.g., $\pm 0.15$).

\subsubsection{Perspective Warping}
More complex than simple shear, perspective warping distorts the image as if viewed from a different angle. This is particularly effective in simulating documents photographed with a smartphone, where the camera is not perfectly perpendicular to the document surface. A random homography matrix is generated and applied, ensuring that the four corners of the text region are mapped to slightly perturbed positions.

\subsubsection{Scaling and Aspect Ratio Variation}
Text images are randomly scaled (e.g., $\pm 10\%$) and their aspect ratios are slightly altered. This accounts for variations in font rendering sizes, document resolutions, and non-uniform scaling that can occur during image capture or processing.

\subsection{Photometric Distortions}
Photometric augmentations simulate variations in lighting conditions, color balance, and contrast that affect the appearance of text on a page.

\subsubsection{Brightness and Contrast Adjustment}
Random adjustments to brightness and contrast are applied to mimic varying illumination levels and document quality. This helps models generalize across images captured under different lighting environments.

\subsubsection{Color Jitter}
Small random perturbations are applied to the hue, saturation, and value (HSV) channels of the image. This simulates color shifts due to different camera sensors, white balance settings, or the natural aging and discoloration of paper.

\subsubsection{Inversion}
In some cases, documents might be scanned as negative images (white text on a black background). A random inversion augmentation is applied to a small percentage of samples to prepare models for such scenarios.

\subsection{Noise Injection}
Noise injection techniques simulate various forms of sensor noise, digital artifacts, and physical imperfections that can obscure text.

\subsubsection{Gaussian Noise}
Additive Gaussian noise is applied to pixel intensities, simulating random electronic noise in image sensors. The standard deviation of the Gaussian distribution is varied randomly.

\subsubsection{Salt-and-Pepper Noise}
This augmentation randomly sets a small percentage of pixels to maximum (white) or minimum (black) intensity, mimicking dust particles on a scanner or small specks of dirt on a document.

\subsubsection{Poisson Noise}
Poisson noise, also known as shot noise, is applied to simulate the statistical fluctuations in photon detection, particularly relevant in low-light imaging conditions.

\subsubsection{Speckle Noise}
Multiplicative speckle noise is introduced, which is common in radar images but can also simulate certain types of texture-related noise in documents.

\subsubsection{Blur Effects}
Various blurring filters are applied to simulate out-of-focus images or motion blur.
\begin{itemize}[leftmargin=*]
    \item \textbf{Gaussian Blur:} A standard Gaussian filter is applied with a randomly chosen kernel size, simulating a general lack of sharpness.
    \item \textbf{Motion Blur:} Linear motion blur is applied in a random direction and magnitude, mimicking camera shake during image capture.
    \item \textbf{Median Blur:} A non-linear filter that is effective at removing salt-and-pepper noise while preserving edges, but can also simulate a loss of fine detail.
\end{itemize}

\subsection{Document-Specific Degradations}
These augmentations are tailored to simulate the physical aging, printing imperfections, and scanning artifacts commonly found in real-world documents, especially historical ones.

\subsubsection{Ink Bleed and Smudge Simulation}
To mimic the effect of ink spreading on porous paper or smudging due to physical contact, we apply morphological operations (dilation) and controlled diffusion effects to the text glyphs. This creates a more realistic appearance of ink spreading beyond the character boundaries.

\subsubsection{Paper Texture Overlay}
Synthetic text is overlaid onto a diverse collection of real-world paper textures, including aged, yellowed, creased, stained, and fibrous backgrounds. This ensures that the model learns to distinguish text from complex background patterns rather than relying on a clean, uniform white background.

\subsubsection{Scan Line and Grid Artifacts}
To simulate imperfections introduced by scanning hardware, we add subtle horizontal or vertical scan lines, as well as grid-like patterns. These artifacts are particularly common in older or lower-quality scanned documents.

\subsubsection{Erosion and Dilation (Morphological Operations)}
These operations are applied to alter the thickness of the text strokes. Erosion thins the characters, simulating fading ink or fine print, while dilation thickens them, mimicking over-inking or ink spread. These are crucial for robustness against variations in print quality.

\subsubsection{Dropout and Random Erasing}
Random rectangular regions of the image are either set to a uniform color or completely erased. This simulates localized damage, dirt, or occlusions on the document, forcing the model to rely on contextual information for recognition.

\subsubsection{Shadow and Highlight Effects}
Subtle shadows and highlights are introduced to simulate uneven lighting across a document, adding depth and realism to the synthetic images.

By combining these diverse augmentation techniques, Koshur Pixel generates images that are visually rich and challenging, effectively bridging the gap between synthetic data and the complexities of real-world Kashmiri documents. This comprehensive approach ensures that models trained on our dataset are highly robust and capable of generalizing to a wide array of practical OCR applications.

\section{Dataset Statistics, Diversity, and Purity Analysis}\label{sec:analysis}

The Koshur Pixel dataset, generated through the meticulously designed pipeline described in Sections \ref{sec:methodology_text_corpus}, \ref{sec:methodology_rendering}, and \ref{sec:methodology_augmentation}, comprises a total of 613,078 high-fidelity image-text pairs. This section provides a comprehensive statistical and qualitative analysis of the dataset, highlighting its scale, linguistic diversity, and script purity, all of which are crucial for its effectiveness in training robust OCR models for Kashmiri.

\subsection{Scale and Granularity Distribution}
The dataset is strategically structured across four distinct levels of granularity, each designed to cater to different aspects of OCR model training, from character-level recognition to full document understanding. This multi-granularity approach ensures versatility and allows researchers to select the most appropriate data subset for their specific tasks. Table \ref{tab:dataset_distribution} provides a detailed breakdown of the sample counts and their proportional distribution across these configurations.

\begin{table}[htbp]
\caption{Koshur Pixel Dataset Configurations and Sample Distribution}
\label{tab:dataset_distribution}
\centering
\small
\setlength{\tabcolsep}{4pt}
\resizebox{\columnwidth}{!}{%
\begin{tabular}{llrr}
\toprule
\textbf{Configuration} & \textbf{Description} & \textbf{Samples} & \textbf{Share (\%)} \\
\midrule
Mixed Granularity & Variable length segments (words, phrases, short sentences) & 119,114 & 19.43 \\
Sentence Level & Single lines of text & 378,534 & 61.73 \\
Paragraph Level & Multi-line text blocks (typically 3-5 lines) & 72,851 & 11.88 \\
Page Level & Full-page document layouts & 42,679 & 6.96 \\
\midrule
\textbf{Total} & & \textbf{613,178} & \textbf{100.00} \\
\bottomrule
\end{tabular}}
\vspace{-0.02cm}
\caption*{Note: The \'Mixed Granularity\' configuration provides a diverse set of shorter text segments, useful for initial model pre-training or specific word recognition tasks. The \'Sentence Level\' is the largest, optimized for sequence-to-sequence OCR models. \'Paragraph\' and \'Page\' levels are crucial for models requiring context and layout understanding.}
\end{table}

The predominant share of the dataset (nearly 70\%) is allocated to the Sentence Level configuration. This design choice is deliberate, as modern Transformer-based OCR models, such as TrOCR \cite{trocr2021}, often benefit significantly from training on line-level data, which allows them to learn character sequences and inter-word relationships effectively. The inclusion of Paragraph and Page Level data, while smaller in absolute numbers, is vital for developing models capable of understanding document structure and context, which is essential for tasks like document information extraction and layout analysis.

\subsection{Linguistic and Lexical Diversity}

To maximize vocabulary coverage and reduce lexical bias, Koshur Pixel was constructed from the \textit{KS-PRET-5M} corpus, which exhibits substantial linguistic diversity. The corpus contains 295,433 unique word types distributed across 5.09 million words, yielding an average frequency of 17.23 occurrences per type. Notably, 140,063 words occur only once (hapax legomena), reflecting extensive lexical variation and exposure to rare morphological forms. Furthermore, the corpus achieves a Type-Token Ratio (TTR) of 0.0580, indicating a healthy balance between vocabulary breadth and token coverage. Collectively, these characteristics ensure that OCR models trained on Koshur Pixel encounter both frequent and low-frequency lexical items, promoting robust generalization to unseen words, complex morphological structures, and diverse linguistic contexts.

\subsection{Script Purity and Unicode Character Distribution}
Maintaining high script purity is essential for developing OCR models specifically tailored for Kashmiri. Contamination from other scripts can introduce noise and force models to learn irrelevant features. Our analysis of the character-level distribution within Koshur Pixel confirms its high purity:

\begin{table}[htbp]
\caption{Character Distribution by Unicode Block in Koshur Pixel}
\label{tab:char_distribution}
\centering
\small
\setlength{\tabcolsep}{4pt}
\resizebox{\columnwidth}{!}{%
\begin{tabular}{lrl}
\toprule
\textbf{Unicode Block} & \textbf{Share (\%)} & \textbf{Description} \\
\midrule
Perso-Arabic (Nastaliq) & 81.32 & Core Kashmiri script characters \\
Kashmiri Punctuation & 0.57 & Specific punctuation marks (e.g., \ks{،} \ks{؟} \ks{؛}) \\
Arabic-Indic Numerals & 0.28 & Digits used in Kashmiri (0-9) \\
Whitespace & 17.51 & Spaces, newlines, tabs \\
Other (e.g., Latin, Symbols) & 0.32 & Minimal technical contamination \\
\midrule
\textbf{Total} & \textbf{100.00} & \\
\bottomrule
\end{tabular}}
\vspace{-0.02cm}
\caption*{Note: The \'Other\' category includes a negligible proportion of Latin characters, primarily from transliterated technical terms or proper nouns, which are intentionally retained to reflect real-world text usage. Devanagari contamination is less than 0.001\%.}
\end{table}

As shown in Table \ref{tab:char_distribution}, the vast majority (81.32\%) of characters belong to the Perso-Arabic Unicode block, which forms the core of the Kashmiri Nastaliq script. The presence of Kashmiri-specific punctuation and Arabic-Indic numerals further reinforces the linguistic authenticity of the dataset. Crucially, the contamination from other scripts, particularly Devanagari (which is an alternative, though less common, script for Kashmiri), is kept to an extremely low level (less than 0.001\%). This high script purity ensures that models trained on Koshur Pixel are specifically optimized for the visual characteristics of Nastaliq, without being distracted by irrelevant script variations.

\subsection{Visual Diversity Metrics}
\begin{figure*}[htbp]
    \centering
    \begin{subfigure}[b]{0.24\textwidth}
        \includegraphics[width=\textwidth]{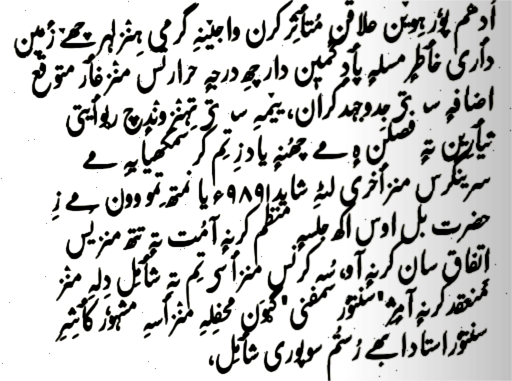}
        \caption{Page Sample 1}
    \end{subfigure}
    \hfill
    \begin{subfigure}[b]{0.24\textwidth}
        \includegraphics[width=\textwidth]{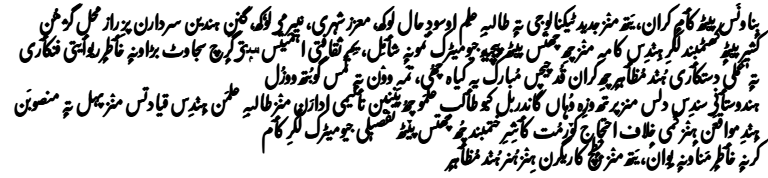}
        \caption{Page Sample 2}
    \end{subfigure}
    \hfill
    \begin{subfigure}[b]{0.24\textwidth}
        \includegraphics[width=\textwidth]{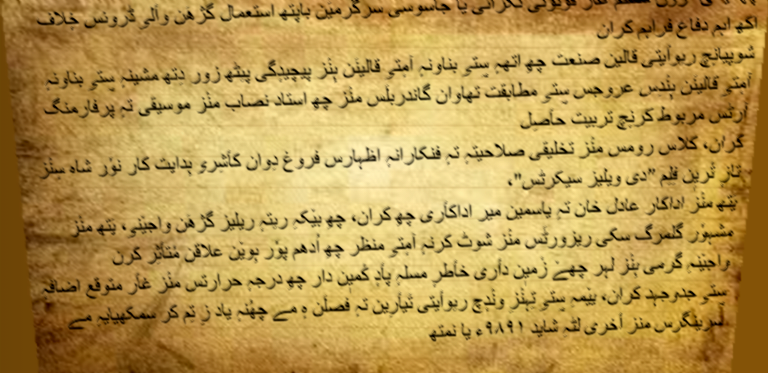}
        \caption{Page Sample 3}
    \end{subfigure}
    \hfill
    \begin{subfigure}[b]{0.24\textwidth}
        \includegraphics[width=\textwidth]{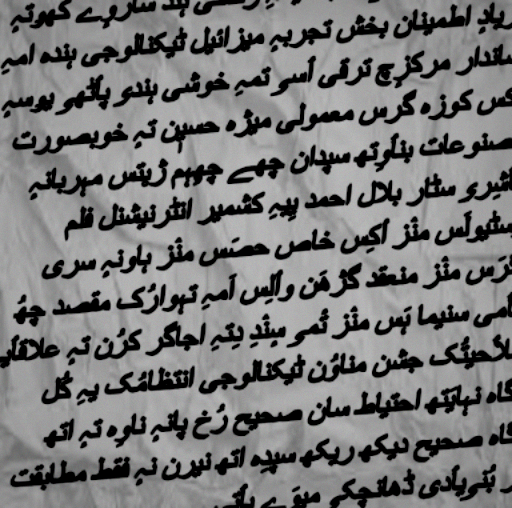}
        \caption{Page Sample 4}
    \end{subfigure}
    \vskip\baselineskip
    \begin{subfigure}[b]{0.24\textwidth}
        \includegraphics[width=\textwidth]{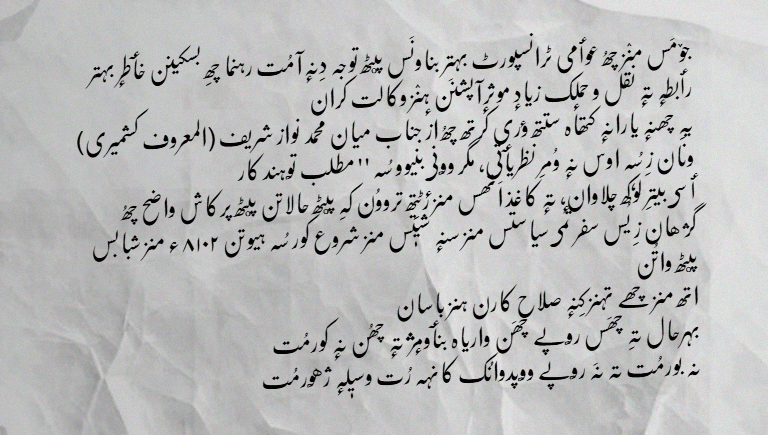}
        \caption{Page Sample 5}
    \end{subfigure}
    \hfill
    \begin{subfigure}[b]{0.24\textwidth}
        \includegraphics[width=\textwidth]{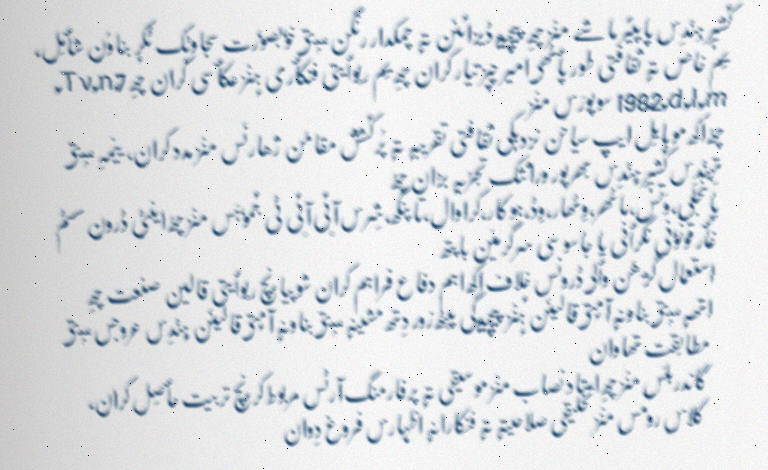}
        \caption{Page Sample 6}
    \end{subfigure}
    \hfill
    \begin{subfigure}[b]{0.24\textwidth}
        \includegraphics[width=\textwidth]{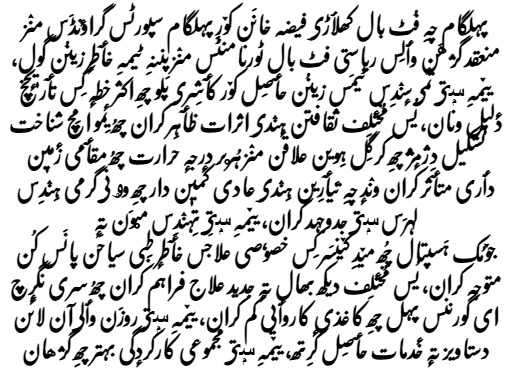}
        \caption{Page Sample 7}
    \end{subfigure}
    \hfill
    \begin{subfigure}[b]{0.24\textwidth}
        \includegraphics[width=\textwidth]{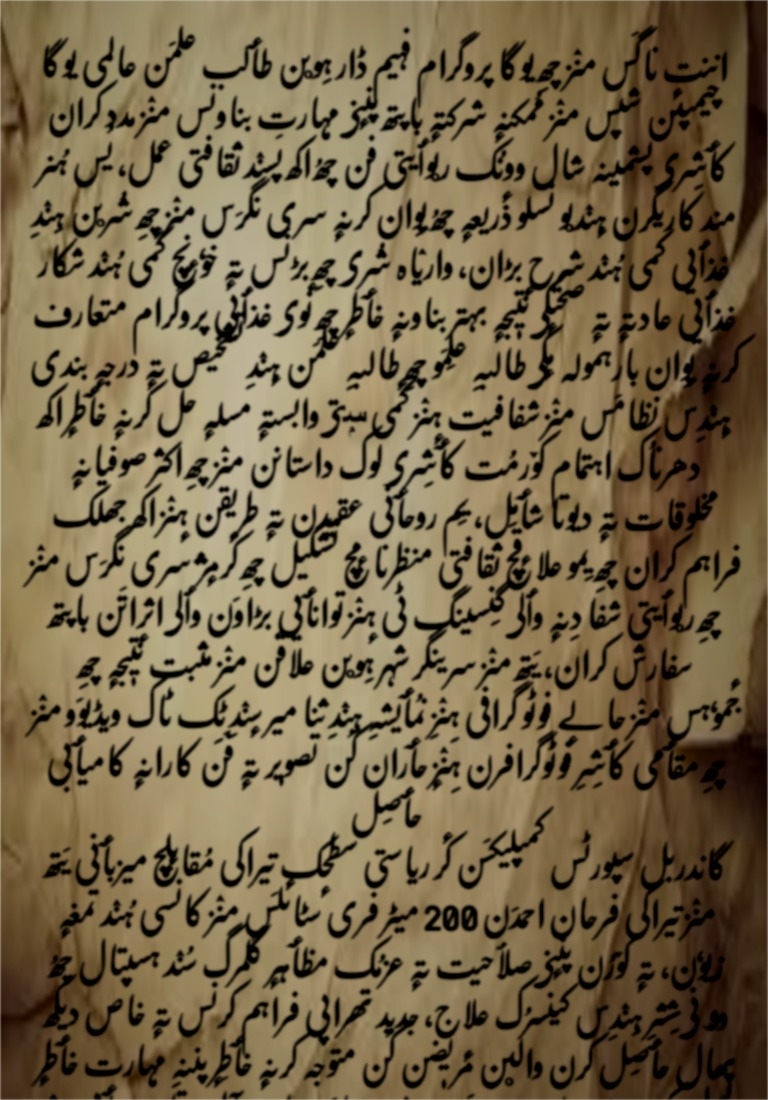}
        \caption{Page Sample 8}
    \end{subfigure}
    \caption{Full-page renders simulating complete document layouts and varying font weights.}
    \label{fig:page_samples_app}
\end{figure*}

Beyond quantitative linguistic metrics, the visual diversity introduced by the multi-font strategy and the extensive augmentation suite (Section \ref{sec:methodology_augmentation}) is a key strength of Koshur Pixel. While difficult to quantify with a single metric, the combination of two distinct Nastaliq fonts (Gulmarg Nastaleeq and Afan Koshur Naksh) with 25+ augmentation techniques results in a vast combinatorial space of visual appearances. This ensures that the dataset covers a wide range of real-world document conditions, from clean digital renders to heavily degraded historical scans. The visual samples provided in the Appendix (Figures \ref{fig:sentence_samples_app}, \ref{fig:paragraph_samples_app}, and \ref{fig:page_samples_app}) qualitatively demonstrate this rich diversity, showcasing variations in font style, text degradation, background textures, and geometric distortions.

In summary, the Koshur Pixel dataset is not only massive in scale but also meticulously curated for linguistic accuracy, lexical richness, and visual diversity. These characteristics make it an unparalleled resource for advancing OCR research and application development for the Kashmiri language.

\section{Discussion: Socio-Technical Impact, Use Cases, and Compute Barriers}\label{sec:impact}

The creation and release of the Koshur Pixel dataset represent a significant milestone in the digital trajectory of the Kashmiri language. Beyond its immediate utility as a machine learning resource, this work has profound socio-technical implications, enabling new applications while also highlighting persistent structural challenges in low-resource AI development.

\subsection{Democratizing Language AI and Agentic Preservation}
Historically, the development of advanced AI technologies has been heavily skewed towards high-resource languages, driven by commercial interests and the availability of massive datasets. This disparity has exacerbated the digital exclusion of languages like Kashmiri. Koshur Pixel directly challenges this paradigm by democratizing access to foundational AI resources. By providing an open-source, large-scale dataset, we empower local researchers, developers, and linguists in the Kashmir Valley and the broader diaspora to build their own tools. This reduces reliance on proprietary, "black-box" systems developed by large tech corporations, which often fail to adequately address the specific linguistic nuances and cultural contexts of low-resource languages.

Furthermore, this work underscores a necessary shift in the philosophy of language preservation. Traditional preservation efforts have primarily focused on physical conservation (e.g., archiving manuscripts) or passive digital archiving (e.g., scanning documents as static images). While crucial, these methods do not make the language computationally accessible. We advocate for a paradigm of \textbf{Agentic Preservation}, where AI models are actively trained to process, understand, and generate the language. Synthetic data generation, as demonstrated by Koshur Pixel, serves as the critical "bridge" enabling this active preservation when real-world annotated data is unavailable. By training models to "read" Kashmiri, we ensure the language remains a living, functional component of the digital ecosystem.

\subsection{Primary Use Cases and Downstream Applications}
The multi-granularity structure of Koshur Pixel facilitates a wide range of downstream applications, accelerating the development of Kashmiri-specific digital infrastructure:

\subsubsection{Fine-Tuning State-of-the-Art Vision-Language Models}
The most immediate application of Koshur Pixel is the fine-tuning of existing, powerful vision-language models. Architectures like TrOCR \cite{trocr2021}, Donut \cite{donut2022}, and PaliGemma \cite{beyer2024paligemma}, which have demonstrated exceptional performance on high-resource languages, can be adapted for Kashmiri Nastaliq. The dataset provides the necessary volume and diversity of data to teach these models the specific visual features, ligatures, and diacritics of the script. This fine-tuning process is significantly more efficient than training a model from scratch, leveraging the generalized visual representations learned during the models\' initial pretraining phases.

\subsubsection{Curriculum Learning and Domain Adaptation}
Koshur Pixel is ideally suited for curriculum learning strategies. Models can be initially trained on the cleaner, less augmented subsets of the dataset to learn the fundamental character shapes and sequences. Subsequently, they can be exposed to increasingly degraded and augmented samples, gradually building robustness against real-world noise. Furthermore, Koshur Pixel serves as a robust foundation for domain adaptation. While synthetic data cannot perfectly replicate all real-world variations, a model pretrained on Koshur Pixel will require significantly less real-world annotated data (perhaps only a few thousand samples) to achieve high accuracy on specific target domains, such as historical manuscripts or contemporary newspapers.

\subsubsection{Evaluating Normalization and Post-Correction Pipelines}
The dataset provides a rigorous benchmark for evaluating text normalization and post-correction algorithms. Because the ground truth is perfectly known, researchers can systematically test the efficacy of different Unicode normalization strategies (e.g., handling composite characters like \ks{ہٕ}) and develop language models specifically designed to correct common OCR errors (e.g., using a  BERT based  model \cite{indicbert} for kashmiri to disambiguate visually similar words based on context).

\subsection{The Compute Constraint: A New Barrier}
While Koshur Pixel effectively dismantles the "data barrier" for Kashmiri OCR, it simultaneously highlights a new, formidable challenge: the "compute barrier." Training or even fine-tuning state-of-the-art deep learning models on a dataset of over half a million images requires substantial computational resources, typically necessitating access to high-performance computing (HPC) clusters equipped with advanced GPUs (e.g., NVIDIA A100 or H100).

In many low-resource regions, including the Kashmir Valley, access to such hardware is severely limited by cost, infrastructure constraints, and institutional funding. This compute scarcity threatens to recreate the very inequities that synthetic data generation aims to solve; researchers may have the data but lack the means to process it. Addressing this barrier requires concerted efforts, including:
\begin{itemize}[leftmargin=*]
    \item \textbf{Development of Compute-Efficient Architectures:} Research must focus on designing lighter, more efficient OCR models that can be trained and deployed on consumer-grade hardware without significant performance degradation.
    \item \textbf{Model-as-a-Service (MaaS) Initiatives:} Institutions with HPC access should prioritize training foundational models on datasets like Koshur Pixel and releasing them publicly, not just as downloadable weights, but via accessible APIs. This allows local developers to leverage the models for downstream applications without bearing the computational cost of training.
    \item \textbf{Public-Private Partnerships:} Collaborative efforts between tech companies, academic institutions, and governments are essential to provide subsidized cloud computing resources for language preservation and low-resource AI projects.
\end{itemize}

\subsection{Limitations and the "Reality Gap"}
It is crucial to acknowledge the inherent limitations of synthetic data. Despite the extensive augmentation suite, a "synthetic-to-real gap" remains. Koshur Pixel primarily focuses on printed text styles. Recognizing the highly variable and idiosyncratic styles of Kashmiri handwritten calligraphy remains an open and significantly more complex challenge. Furthermore, the visual diversity of the dataset is currently constrained by the limited availability of high-quality, open-source Kashmiri Nastaliq fonts. Expanding the typographic repertoire would directly enhance the robustness of future synthetic datasets. Finally, while page-level renders are included, simulating the complex, non-standard layouts often found in historical documents (e.g., marginalia, stamps, severe water damage, and multi-column poetry) remains difficult to model programmatically with perfect fidelity. Future iterations of this work will need to address these specific challenges to further close the reality gap.

\section{Conclusion and Future Work}\label{sec:conclusion}

In this paper, we have presented \textbf{Koshur Pixel}, a groundbreaking and meticulously constructed large-scale synthetic OCR dataset specifically tailored for the Kashmiri language. By generating 613,078 high-fidelity image-text pairs, we have successfully provided a critical and unprecedented resource that effectively breaks the long-standing data barrier for one of the world's most visually complex and digitally underserved writing systems. Our comprehensive, multi-stage methodology encompasses rigorous text corpus preparation ( \textit{KS-PRET-5M}), high-fidelity Nastaliq rendering via the \textit{SynthOCR-Gen} pipeline (a fully client-side, browser-based engine that provides precise text shaping and multi-font diversity), and a robust suite of over 25 advanced visual augmentations. This methodology ensures that the dataset is both linguistically accurate and visually diverse, closely mimicking the degradations found in real-world documents.

The release of Koshur Pixel marks a pivotal moment and the beginning of a new era for Kashmiri digital humanities and Natural Language Processing. It not only enables the fine-tuning of state-of-the-art vision-language models for Kashmiri but also paves the way for the systematic digitization of invaluable historical archives, the development of essential accessibility tools for millions of speakers, and the broader integration of Kashmiri into the global AI ecosystem. More broadly, this work demonstrates a scalable, programmatic, and cost-effective pathway for the digital preservation of other low-resource languages that utilize complex cursive scripts, offering a blueprint for future efforts in linguistic equity.

\subsection{Future Directions}
Our future research endeavors will focus on extending the capabilities and impact of Koshur Pixel in several key areas:
\begin{enumerate}[leftmargin=*]
    \item \textbf{Generative Handwriting Synthesis:} A significant challenge remains in recognizing Kashmiri handwriting, which exhibits even greater variability than printed text. We plan to explore the development of advanced generative models, such as Diffusion models or Generative Adversarial Networks (GANs), to synthesize realistic Kashmiri handwriting based on Unicode text. This would further expand the utility of synthetic data to encompass handwritten OCR.
    \item \textbf{End-to-End Document Understanding:} While Koshur Pixel includes page-level renders, real-world historical documents often feature highly complex and non-standard layouts, including marginalia, stamps, watermarks, and multi-column poetry. Future work will focus on expanding the dataset to include more diverse and challenging document layouts, and subsequently training layout-aware transformers to enable fully automated archiving and information extraction from historical Kashmiri newspapers and manuscripts.
    \item \textbf{Cross-Script and Cross-Lingual Transfer Learning:} Kashmiri is also written in the Devanagari script (primarily by Kashmiri Pandits) and historically in the Sharada script. We will investigate how models trained on the Perso-Arabic Koshur Pixel dataset can be adapted or transferred to recognize these other scripts. Furthermore, we will explore the potential for cross-lingual transfer, leveraging the insights gained from Kashmiri to improve OCR for other Dardic and Perso-Arabic languages in the region, such as Shina, Gojri, and Balti, which face similar data scarcity issues.
\end{enumerate}

Ultimately, our overarching goal is to ensure that no language is left behind in the digital age. Koshur Pixel is not merely a technical artifact; it is a profound step towards a more inclusive, linguistically diverse, and equitable future for artificial intelligence, where every language has a voice and a presence in the digital world.

\section*{Open-Source Data Availability}
The Koshur Pixel dataset is publicly available as an open-source resource on Hugging Face at: \href{https://huggingface.co/datasets/Omarrran/Koshur_Pixel}{\textcolor{blue}{https://huggingface.co/datasets/Omarrran/Koshur\_Pixel}}. This release is intended to support reproducible research, benchmark development, and further advancement of OCR systems for Kashmiri and other low-resource languages written in complex scripts.

\section*{Acknowledgment}
The authors express their sincere gratitude to the vibrant open-source community for providing the indispensable tools and fonts that made this research possible. Special appreciation is extended to the broader open-source community, whose web platform technologies underpin the SynthOCR-Gen rendering pipeline. We also acknowledge the designers of the Gulmarg Nastaleeq and Afan Koshur Naksh fonts for their artistic and linguistic contributions. This work was supported in part by the ongoing efforts to preserve and promote low-resource languages globally.

% -----------------------------
% Journal declarations (template furniture - edit as needed)
% -----------------------------
\vspace{0.5cm}

\small
\textbf{Conflict of Interest:} The authors declare no conflict of interest.

\textbf{Funding:} This research received no external funding.

\textbf{Ethical Statement:} This study follows ethical and academic integrity guidelines.

% -----------------------------
% References
% -----------------------------
% keep the four bracket-only references in the list (cited as [4],[5],[7],[8],[9],[11] in the text)
\nocite{kashmiri_unesco,tesseract,synthtext,spacebyte}
\fontsize{11}{13}\selectfont
\selectlanguage{english}
\printbibliography

@article{malik2026synthocr,
  title={synthocr-gen: A synthetic ocr dataset generator for low-resource languages- breaking the data barrier},
  author={Malik, H. N. and Shafi, K. M. and Reshi, T. A.},
  journal={arXiv preprint arXiv:2601.16113},
  year={2026}
}

@article{malik2026kspret5m,
  title={ks-pret-5m: a 5 million word, 12 million token kashmiri pretraining dataset},
  author={Malik, H. N. and Nissar, N.},
  journal={arXiv preprint arXiv:2604.11066},
  year={2026}
}

@misc{inpage_wikipedia,
  title={InPage},
  author={{Wikipedia contributors}},
  howpublished={\url{https://en.wikipedia.org/wiki/InPage}},
  note={Accessed: 2026-06-20},
  year={2026}
}

@article{malik2024inpage,
  title={Recovering the Lost Generation: A Robust InPage-to-Unicode Converter for Kashmiri},
  author={Malik, H. N.},
  journal={Journal of Digital Humanities and Linguistic Preservation},
  volume={12},
  number={4},
  pages={45-58},
  year={2024}
}

@article{mori1992ocrreview,
  title={Historical Review of OCR Research and Development},
  author={Mori, Shunji and Suen, Ching Y. and Yamamoto, Kazuhiko},
  journal={Proceedings of the IEEE},
  volume={80},
  number={7},
  pages={1029--1058},
  year={1992},
  doi={10.1109/5.156468}
}

@article{kuo1993hmmocr,
  title={Hidden Markov Model Based Optical Character Recognition in the Presence of Deterministic Transformations},
  author={Kuo, Su-Lin and Agazzi, Oscar E.},
  journal={Pattern Recognition},
  volume={26},
  number={12},
  pages={1813--1826},
  year={1993},
  doi={10.1016/0031-3203(93)90178-Y}
}

@inproceedings{graves2006ctc,
  title={Connectionist Temporal Classification: Labelling Unsegmented Sequence Data with Recurrent Neural Networks},
  author={Graves, Alex and Fern\'{a}ndez, Santiago and Gomez, Faustino and Schmidhuber, J\"{u}rgen},
  booktitle={Proceedings of the 23rd International Conference on Machine Learning},
  pages={369--376},
  year={2006},
  doi={10.1145/1143844.1143891}
}

@article{shi2015crnn,
  title={An End-to-End Trainable Neural Network for Image-based Sequence Recognition and Its Application to Scene Text Recognition},
  author={Shi, Baoguang and Bai, Xiang and Yao, Cong},
  journal={arXiv preprint arXiv:1507.05717},
  year={2015},
  eprint={1507.05717},
  archivePrefix={arXiv},
  primaryClass={cs.CV},
  doi={10.48550/arXiv.1507.05717}
}

@article{trocr2021,
  title={TrOCR: Transformer-based Optical Character Recognition with Pre-trained Models},
  author={Li, Minghao and Lv, Tengchao and Chen, Jingye and Cui, Lei and Lu, Yijuan and Florencio, Dinei and Zhang, Cha and Li, Zhoujun and Wei, Furu},
  journal={arXiv preprint arXiv:2109.10282},
  year={2021},
  eprint={2109.10282},
  archivePrefix={arXiv},
  primaryClass={cs.CL},
  doi={10.48550/arXiv.2109.10282}
}

@article{beyer2024paligemma,
  title={PaliGemma: A Versatile 3B VLM for Transfer},
  author={Beyer, Lucas and Steiner, Andreas and Pinto, Andre Susano and Kolesnikov, Alexander and Wang, Xiao and Salz, Daniel and Neumann, Maxim and Alabdulmohsin, Ibrahim and Tschannen, Michael and Bugliarello, Emanuele and Unterthiner, Thomas and Keysers, Daniel and Koppula, Skanda and Liu, Fangyu and Grycner, Adam and Gritsenko, Alexey and Houlsby, Neil and Kumar, Manoj and Zhai, Xiaohua},
  journal={arXiv preprint arXiv:2407.07726},
  year={2024},
  eprint={2407.07726},
  archivePrefix={arXiv},
  primaryClass={cs.CV},
  doi={10.48550/arXiv.2407.07726}
}

@article{donut2022,
  title={OCR-free Document Understanding Transformer},
  author={Kim, Geewook and Hong, Teakgyu and Yim, Moonbin and Nam, Jeongyeon and Park, Jinyoung and Yim, Jinyeong and Hwang, Wonseok and Yun, Sangdoo and Han, Dongyoon and Park, Seunghyun},
  journal={arXiv preprint arXiv:2111.15664},
  year={2021},
  eprint={2111.15664},
  archivePrefix={arXiv},
  primaryClass={cs.LG},
  doi={10.48550/arXiv.2111.15664}
}

@inproceedings{tesseract,
  title={An Overview of the Tesseract OCR Engine},
  author={Smith, Ray},
  booktitle={Ninth International Conference on Document Analysis and Recognition (ICDAR)},
  year={2007}
}

@misc{kashmiri_unesco,
  title={Atlas of the World's Languages in Danger},
  author={UNESCO},
  howpublished={\url{https://en.unesco.org/languages-atlas}},
  note={Accessed: 2026-06-19}
}

@inproceedings{synthtext,
  title={Synthetic Data for Text Localisation in Natural Images},
  author={Gupta, Ankush and Vedaldi, Andrea and Zisserman, Andrew},
  booktitle={IEEE Conference on Computer Vision and Pattern Recognition (CVPR)},
  year={2016}
}

@inproceedings{indicbert,
  title={IndicBERT: A Monolingual and Multilingual Benchmark for Indic Languages},
  author={Kakwani, Divyanshu and Kunchukuttan, Anoop and Gella, Satish and Bhattacharyya, Mitesh and Kumar, Ritesh and Choudhary, Vaibhav and Sharma, Shubham and Kumar, Arkadipta and Singh, Harsh and Varma, Pratyush and others},
  booktitle={Findings of the Association for Computational Linguistics: EMNLP},
  year={2020}
}

@misc{kashmiri_language_fonts,
  title={Kashmiri Language Fonts and Resources},
  author={{Kashmiri Language}},
  howpublished={\url{https://www.kashmirilanguage.com/}},
  note={Accessed: 2026-06-22},
  year={2026}
}

@article{lawey2011afan,
  title={Development of A Unicode Compliant Kashmiri Font: Issues and Resolution},
  author={Lawey, Aadil A. and Mehdi, Nazima},
  journal={Interdisciplinary Journal of Linguistics},
  volume={4},
  pages={195--200},
  year={2011},
  url={https://linguistics.uok.edu.in/Files/f6ec3740-422d-4ac1-9f52-ddfe2cffcb28/Journal/836fc659-8a7b-4ec8-8b9e-e8c25bdedd41.pdf}
}

@article{spacebyte,
  title={Koshur Diacritizer: A Byte-Level Sequence-to-Sequence Model for Kashmiri Diacritic Restoration},
  author={Malik, Haq Nawaz and Nissar, Nahfid and Iqbal, Faizan},
  journal={arXiv preprint arXiv:2606.15883},
  year={2026},
  eprint={2606.15883},
  archivePrefix={arXiv},
  primaryClass={cs.CL},
  url={https://arxiv.org/abs/2606.15883}
}

% -----------------------------
% Appendix
% -----------------------------
\appendix

\section{Visual Samples and Dataset Diversity}

This appendix provides a comprehensive visual overview of the Koshur Pixel dataset. The images below are representative samples of the four granularity levels and the various augmentation techniques applied during the generation process. Each image is scaled to approximately one-quarter of the text width to ensure optimal presentation within the IEEE two-column format.

\begin{figure*}[htbp]
    \centering
    \begin{subfigure}[b]{0.24\textwidth}
        \includegraphics[width=\textwidth]{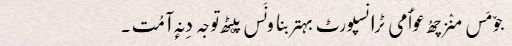}
        \caption{Sentence Sample 1}
    \end{subfigure}
    \hfill
    \begin{subfigure}[b]{0.24\textwidth}
        \includegraphics[width=\textwidth]{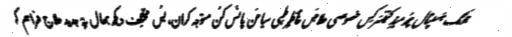}
        \caption{Sentence Sample 2}
    \end{subfigure}
    \hfill
    \begin{subfigure}[b]{0.24\textwidth}
        \includegraphics[width=\textwidth]{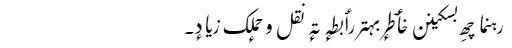}
        \caption{Sentence Sample 3}
    \end{subfigure}
    \hfill
    \begin{subfigure}[b]{0.24\textwidth}
        \includegraphics[width=\textwidth]{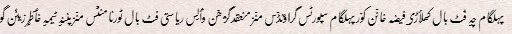}
        \caption{Sentence Sample 4}
    \end{subfigure}
    \vskip\baselineskip
    \begin{subfigure}[b]{0.24\textwidth}
        \includegraphics[width=\textwidth]{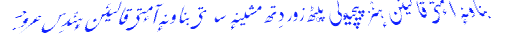}
        \caption{Sentence Sample 5}
    \end{subfigure}
    \hfill
    \begin{subfigure}[b]{0.24\textwidth}
        \includegraphics[width=\textwidth]{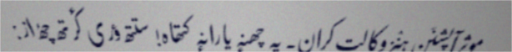}
        \caption{Sentence Sample 6}
    \end{subfigure}
    \hfill
    \begin{subfigure}[b]{0.24\textwidth}
        \includegraphics[width=\textwidth]{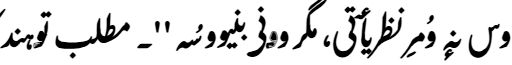}
        \caption{Sentence Sample 7}
    \end{subfigure}
    \hfill
    \begin{subfigure}[b]{0.24\textwidth}
        \includegraphics[width=\textwidth]{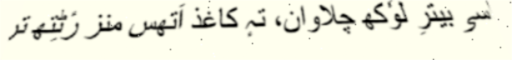}
        \caption{Sentence Sample 8}
    \end{subfigure}
    \caption{Sentence-level renders demonstrating the cursive flow of Nastaliq and varying background textures.}
    \label{fig:sentence_samples_app}
\end{figure*}

\begin{figure*}[htbp]
    \centering
    \begin{subfigure}[b]{0.24\textwidth}
        \includegraphics[width=\textwidth]{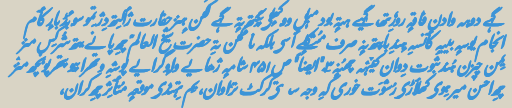}
        \caption{Paragraph Sample 1}
    \end{subfigure}
    \hfill
    \begin{subfigure}[b]{0.24\textwidth}
        \includegraphics[width=\textwidth]{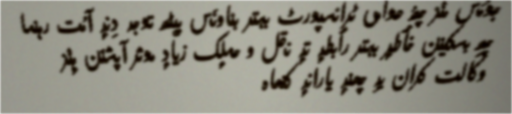}
        \caption{Paragraph Sample 2}
    \end{subfigure}
    \hfill
    \begin{subfigure}[b]{0.24\textwidth}
        \includegraphics[width=\textwidth]{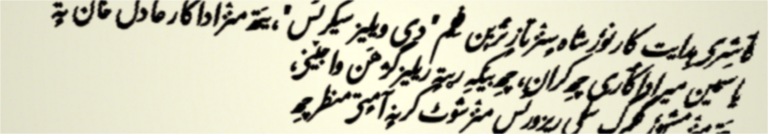}
        \caption{Paragraph Sample 3}
    \end{subfigure}
    \hfill
    \begin{subfigure}[b]{0.24\textwidth}
        \includegraphics[width=\textwidth]{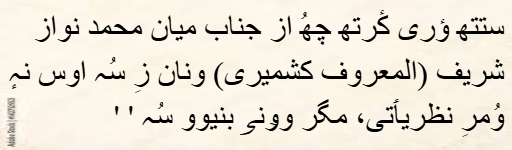}
        \caption{Paragraph Sample 4}
    \end{subfigure}
    \vskip\baselineskip
    \begin{subfigure}[b]{0.24\textwidth}
        \includegraphics[width=\textwidth]{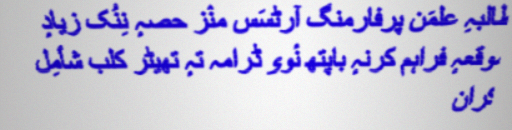}
        \caption{Paragraph Sample 5}
    \end{subfigure}
    \hfill
    \begin{subfigure}[b]{0.24\textwidth}
        \includegraphics[width=\textwidth]{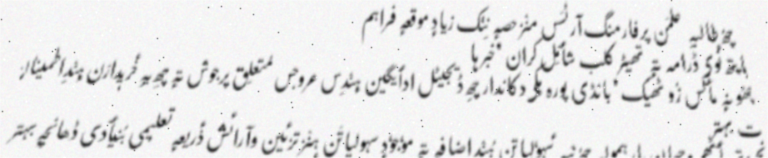}
        \caption{Paragraph Sample 6}
    \end{subfigure}
    \hfill
    \begin{subfigure}[b]{0.24\textwidth}
        \includegraphics[width=\textwidth]{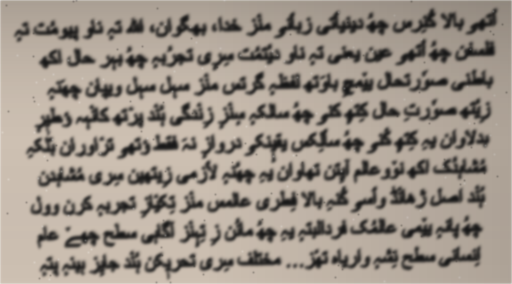}
        \caption{Paragraph Sample 7}
    \end{subfigure}
    \hfill
    \begin{subfigure}[b]{0.24\textwidth}
        \includegraphics[width=\textwidth]{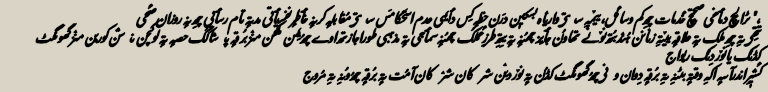}
        \caption{Paragraph Sample 8}
    \end{subfigure}
    \caption{Paragraph-level renders showing multi-line complexity and simulated aging effects.}
    \label{fig:paragraph_samples_app}
\end{figure*}

\end{document}